# Beyond the Keyword: A Human-Annotated Benchmark for Arabic Anxiety-Related Discourse

**Shimaa Ibrahim[1], Mabrouka Bessghaier[1], Md. Rafiul Biswas[2], Insaf Elouard[1], Wajdi Zaghouani[1]**

[1]Northwestern University in Qatar, Doha, Qatar
[2]Hamad Bin Khalifa University, Doha, Qatar
{shimaa.ibrahim, mabrouka.bessghaier, wajdi.zaghouani}@northwestern.edu,
insafelouard2029@u.northwestern.edu, mbiswas@hbku.edu.qa

## Abstract

Keyword retrieval returns expressions of personal anxiety together with many other uses of anxiety-related language. We present a benchmark of 1,000 Arabic Twitter/X posts retrieved with قلق (qalaq, "anxiety/worry") and annotated for six discourse functions. Two native Arabicspeaking annotators labeled every post independently, reaching 51.1% raw agreement (Cohen's $\kappa$=0.378). We then reviewed all 489 disagreements and all 511 agreements under the same guidelines. The review changed 120 of the agreed labels (23.5%), which shows that agreements need auditing as well as disagreements. The final labels form a consensus reference; on a random sample of 200 posts, an additional annotator agreed with it on 60.5% of posts ($\kappa$=0.517). Against this reference, FanarC-2-27B reaches 79.8% accuracy and macro-F1 0.784 ($\kappa$=0.749). Its errors concentrate in the residual category (41.1% of errors; precision 0.905, recall 0.447), whose definition is internally inconsistent in the executed prompt, so the low recall cannot be attributed to the model alone. Sociopolitical Anxiety (27.1%) outnumbers Personal Emotional Expression (22.0%); these are assignment proportions in the sample, not estimates of prevalence or retrieval precision. We release the guidelines, a residualsubtype scheme, a fixed data split, label provenance, and evaluation code.

## 1 Introduction

Arabic mental health NLP frequently starts from keyword-retrieved corpora: a term is chosen, posts containing it are collected, and the collection is treated as relevant to the term's topic. Labels are then applied by lexicon matching, classifiers, or large language models (LLMs). Keyword matching, however, guarantees neither that the retrieved posts describe the intended phenomenon nor that annotators can categorize them consistently.

The Arabic term قلق (qalaq, "anxiety/worry") illustrates the problem. It can describe a personal experience of anxiety (انا قلقان من بكرة, "I am anxious about tomorrow"), a diplomatic or journalistic formula (أعرب عن قلقه البالغ, "expressed grave concern"), or general concern about a situation (الأرقام مقلقة للسوق, "the figures are worrying for the market"). A collection retrieved with قلق therefore cannot be assumed to consist of personal expressions of anxiety without evidence.

We drew 1,000 posts from such a collection, had two annotators label each one independently for discourse function, built a reference label for every post through a consensus review under the same guidelines, and evaluated Fanar-C-2-27B against that reference. The reference was finalized without accesstothe model'spredictions. First-roundagreement was low, and the guidelines contained overlapping instructions for the residual category and Sociopolitical Anxiety. We report the first-round agreement and its disagreement patterns separately from the review that produced the final reference. We also compare an additional annotator's labels with the final reference on a random sample of 200 posts (Section 5.5). Fanar reaches F1 between 0.77 and 0.87 on all five substantive categories but misses more than half of the residual posts. Because first-round agreement compares two independent annotators while model agreement compares the model with a consensus reference, the two figures are not directly comparable. **Contributions.**

1. A benchmark of 1,000 Arabic قلق posts labeled for six discourse functions, with reference labels built by reviewing all 489 first-round disagreements and all 511 agreements, and with per-post provenance recording how each reference label relates to the two initial annotations.

2. A reliability analysis of first-round agreement ($\kappa$=0.378) that localizes disagreement by category, together with an audit showing that 23.5% of initially agreed labels did not survive review, and a 200-post comparison of the final reference with an additional annotator.

3. An evaluation of Fanar-C-2-27B with bootstrap confidence intervals, two reference baselines, and an error analysis showing that the residual category accounts for the largest share of errors.

4. An operational decision scheme for the residual category that separates its subtypes, and a fixed stratified split for future supervised baselines.

## 2 Related Work

### 2.1 Arabic Mental Health NLP and Social Media Resources

Arabicmental health NLP hasgrownin recentyears, but most work addresses supervised classification of depression, suicidal ideation, or mental health questions, often with limited demographic reporting (Alasmari, 2025). Resources include CairoDep for depression detection (El-Ramly et al., 2021), AraDepSu for depression and suicidal ideation (Hassib et al., 2022), MentalQA for mental health questionanswering(Alhuzalietal.,2024), CARMA for multi-condition annotation on Reddit (Mankarious and Zirikly, 2025), and mood-focused resources (Maghraby and Ali, 2022). MentalMARBERT (Almalki et al., 2026) builds a 50,670-tweet corpus labeled for six disorder categories with Krippendorff's $\alpha$=0.733 and reports domain-adapted MARBERT baselines. That work and ours answer different questions. MentalMARBERT asks which clinical category a post signals; we ask what communicative function a single polysemous keyword serves, and five of our six categories carry no clinical meaning. Our benchmark is smaller and its first-round agreement is lower, which we treat as a finding about the task rather than something to hide.

Arabic NLP also offers many social media resources relevant to discourse analysis: multidialectal stance corpora (Charfi et al., 2024), hate speech datasets (Zaghouani et al., 2024), infodemic shared tasks (Shaar et al., 2021; Alam et al., 2021), emotion and hope speech resources (Zaghouani and Biswas, 2025), monitoring platforms such as MARSAD (Biswas et al., 2025), and multidimensional resources in which stance, sentiment and emotion are annotated independently (Laabar and Zaghouani, 2024). Our scheme instead assigns one primary function per post through a tie-break order, so category proportions cannot reflect every function a post serves (Section 7).

**2.2 Mental Health Discourse, Benchmarks, and LLM Annotation** Work on mental health discourse on Twitter/X finds that emotional framing and interaction patterns matter more than attention-seeking for how widely messages circulate (Pavlova and Berkers, 2020, 2022). NLP studies have improved the detection of mental health signals with feature-based models, graph-based emotion representations, and instruction-tuned LLMs (Zhang et al., 2022; Mao and Han, 2025; Yang et al., 2024); related studies examine the romanticization of mental illness (Issaka et al., 2024), digital media use and adolescent well-being (Svensson et al., 2022), and social anxiety and emoji use (Bendl et al., 2025). MentaLLaMA generates rationales alongside predictions (Yang et al., 2024); whether rationales help separate readable out-of-scope posts from substantive ones is an open question for our residual category.

Chhetri et al. (2026) build mental health benchmarks on Bluesky with lexicon filtering, LLMassisted labels, and human-adjudicated validation subsets, and find that weak-label failure modes differ by task. We share their emphasis on label provenance and release a per-post record of how each reference label was reached (Section 8).

On the modeling side, zero-shot LLMs have been found to trail smaller fine-tuned transformers on computational social science classification, with prompt design shifting accuracy by more than ten points (Mu et al., 2023). Alternative prompting paradigms such as noisy-channel prompting improve few-shot stability (Min et al., 2022). We therefore present our LLM result as a characterization of one model under one prompt, not as evidence that LLMs match or exceed fine-tuned classifiers on this task.

**2.3 Framing and Arabic Pragmatics** The same event can be framed differently depending on social position and communicative intent: Shurafa et al. (2020) describe divergent framing of COVID-19 by supporters of U.S. political parties, and Shestakov and Zaghouani (2024) analyze digital framing of the Sheikh Jarrah evictions. During regional crises, a term that names personal anxiety is also routinely used for collective political concern. Because the keyword alone cannot separate these uses, we classify discourse function rather than treat keyword presence as a proxy for psychological content.

## 3 Data Collection and Sampling

### 3.1 Collection

Arabic posts were collected from Twitter/X through Meltwater with a query on قلق and an Arabic language filter, covering 2024-06-05 to 2024-12-31 (210 days). Meltwater caps each export at 20,000 items. Each item carries the post text, timestamps, content type (original, reply, quote, repost), and visibility fields including Reach, an audience estimate based on followers rather than impressions. Two text fields exist: Opening Text, truncated after 144 characters, and Hit Sentence, up to 282 characters. We used Hit Sentence throughout. Using Opening Text would have removed a median of 68 characters from one half of the collection and 2 characters from the other, which matters for any lengthdependent measurement. Truncation in provider exports is a recurring and under-documented problem in social media corpus construction.

### 3.2 Sampling frame

The collection was retrieved as two exports of 20,000 items each, split by Meltwater's inferred author gender. This field was used only to stay within the export cap; it was not used in annotation or analysis. It nevertheless determined which posts entered the pool, so the corpus is not a simple random sample of قلق posts in the period, and every claim in this paper refers to this sampling frame.

Exact-duplicate texts (47% of the collection) were removed. We then drew 1,000 posts at random from the deduplicated pool, stratified by Content Type to match the content-type proportions of the collection before deduplication. We made this choice because deduplication removes reposts far more often than original posts, and the visible repost text is part of what users encounter. The consequence is that the sample over-represents reposted content relative to the deduplicated pool, and its category proportions should not be read as properties of either the platform or the full collection. We did not validate the sample against an independent random draw.

The sample contains 338 original posts, 309 reposts, 276 replies and 77 quotes, with a median length of 156 characters. Two notions of repost coexist: Meltwater's Content Type marks 309 posts as reposts, while 252 posts carry a literal text prefix (207 RT @, 45 QT @). The visibility analysis (Appendix E) uses Content Type. Annotators labeled the visible text as presented and did not retrieve original posts, so human and model annotation units are identical.

**Dialect and register.** Dialect and register were not annotated, and we do not report a distribution. The examples in this paper include Modern Standard Arabic news formulas and Egyptian and Gulf colloquial forms, but we make no claim about their proportions.

## 4 Annotation Scheme

Each post receives one of six labels:

- **Personal Emotional Expression**: the author expresses their own anxiety, distress or worry, in the first person, about their own situation.

- **Sociopolitical Anxiety**: worry or concern about politics, war, the economy or a collective situation, whether felt by the author, attributed to others, or reported.

- **Philosophical / Advisory**: advice, reflection or general wisdom about anxiety, addressed

to others rather than describing the author's current state.

- **Religious / Spiritual Coping**: prayer, supplication (دعاء), a Quranic verse, a hadith, or another invocation of God in connection with anxiety.
- **Humorous / Ironic Use**: jokes, memes, sarcasm or mockery involving anxiety.
- **Not interpretable** (residual): any post that fits none of the five categories above. Despite its name, this label covers readable posts that are out of scope (such as advertisements) as well as fragments and genuinely uninterpretable text.

The codebook gives one example per category, a tie-break order for posts that fit several categories (humour, then religious framing, then outward advice, then the author's own state, then public events, then residual), and rules for reposts, advertisements and truncated text (Appendix B).

| Category | A1/A2 | $\kappa$ |
|---|---|---|
| Personal Emotional Expr. | 216 / 214 | 0.419 |
| Sociopolitical Anxiety | 120 / 229 | 0.371 |
| Philosophical / Advisory | 144 / 127 | 0.347 |
| Religious / Spiritual | 68 / 48 | 0.470 |
| Humorous / Ironic | 72 / 86 | 0.396 |
| Not interpretable | 380 / 296 | 0.339 |
| **Overall** | **0.378** [0.338, 0.417] | |
| *Raw agreement* | | *51.1%* |

Table 1: First-round inter-annotator agreement, $n$=1000. Per-category $\kappa$ is one-vs-rest. Counts show how often each annotator used each category.

**Overlapping instructions.** The guidelines contained two overlaps. First, the residual rules mentioned news reports, while Sociopolitical Anxiety covered concern about political events. Second, the residual category held both readable out-of-scope content and uninterpretable content. During review we applied the existing definitions and tie-break rules and resolved differences by consensus; the codebook was not revised.

**Residual subtypes.** To make the residual category usable for future annotation and error analysis, Appendix C defines four subtypes with an ordered decision procedure: *R1 advertisement*, *R2 readable but out of scope*, *R3 insufficient context*, and *R4 uninterpretable*. These subtypes correspond to the sub-reasons recorded during our review of the agreed posts. They were not part of the firstround guidelines, and we report subtype counts only where they were recorded.

# 5 Annotation Reliability

## 5.1 Procedure

Two native Arabic-speaking annotators labeled all 1,000 posts independently, blind to each other's labels and without access to model output. The reference labels were also finalized without access to the model's predictions or results.

## 5.2 First-round agreement

Table 1 reports first-round agreement. Raw agreement was 51.1% and Cohen's $\kappa$ was 0.378 (95% CI [0.338, 0.417]), *fair* under Landis and Koch (1977). All intervals in this paper are percentile bootstrap intervals with $B$=2000 resamples and seed 42.

## 5.3 Disagreement is concentrated, not diffuse

The residual category appears in the three largest confusion cells, paired with Sociopolitical Anxiety (93 posts), Personal Emotional Expression (89) and Philosophical / Advisory (62). The largest marginal gap between annotators is in Sociopolitical Anxiety (120 vs. 229), followed by the residual category (380 vs. 296); both are involved in the news-report overlap described in Section 4.

| Variant | $n$ | $\kappa$ |
|---|---|---|
| As administered | 1000 | 0.378 |
| Merging Sociopol. + residual | 1000 | 0.403 |
| Excluding residual category | 513 | 0.514 |

Table 2: Diagnostic decomposition of first-round agreement. These variants localize disagreement; they are not estimates of agreement under any revised guideline.

Table 2 shows agreement under alternative treatments of the residual category.

MergingSociopolitical Anxiety with the residual category raises $\kappa$ from 0.378to0.403. Restrictingtothe513poststhatboth annotators placed in substantive categories raises $\kappa$ to 0.514.

Per-category one-vs-rest $\kappa$ nonetheless stays between 0.339 and 0.470, so the residual category is the least reliable boundary but not the only one.

### 5.4 Building the reference

**Disagreements.** After the first round, the first author reviewed all 1,000 posts against the guidelines and proposed a reference label for each. A second author checked every proposal, and differences were resolved by consensus. Both reviewers are native Arabic speakers. For the 489 disagreements, the referencelabel matched the first annotator in 164 cases (33.5%), the second annotator in 245 (50.1%), and neither in 80 (16.4%). No reason codes were recorded for these 489 decisions, so Section 6.3 examines label movement rather than its causes.

**Agreements.** Two annotators agreeing does not guarantee that a label follows the guidelines, and before this step 189 of the 244 posts then labeled residual came from unreviewed agreements. We therefore reviewed all 511 agreed posts in the same way. The review changed 120 labels (23.5%). Ninety-six changes involved posts agreed as residual, so the category retained 49.2% of its 189 agreed posts; 57 of the 96 moved to Sociopolitical Anxiety under the news-report rule. The other 24 changes were spread across Personal Emotional Expression (18 changes; 84.6% retained), Sociopolitical Anxiety (4; 95.1%) and Religious / Spiritual Coping (2; 93.1%). Humorous / Ironic Use and Philosophical / Advisory were unchanged. These 24 changes may reflect annotation errors, reinterpretation of categories, or tie-break application; the transitions alone cannot distinguish among these explanations.

**What the reference is and is not.** All 1,000 reference labels were reviewed under one codebook. The reference is a consensus reference, not an independently replicated gold standard. Because the same two people reviewed every post, anchoring on the proposed labels and gradual drift in how boundary rules were applied are possible. The first-round $\kappa$ remains the only measure of agreement between two independent annotators. Section 5.5 compares one additional annotator with the final reference on a random sample. A larger independent annotation using the residual subtypes in Appendix C would provide further evidence about the reliability of the final labels.

### 5.5 Validation sample

After the reference was finalized, an additional annotator labeled a simple random sample of 200 of the1,000posts. Thesamplewasselectedwithoutreplacement using NumPy RandomState with seed 42 and was not stratified. The annotator received the post texts and written guidelines, including the six category definitions, the tie-break order and the additional rules in Appendix B. These included an explicit rule assigning news reports and diplomatic expressions of concern to Sociopolitical Anxiety. The annotator did not have access to the first-round labels, the reference labels or the model predictions.

The annotator agreed with the reference on 121 posts (60.5%, 95% CI [54.0, 67.5]), with Cohen's $\kappa$ of 0.517 (95% CI [0.437, 0.598]). Agreement was 75.5% (80 of 106) on sampled posts whose reference labels were recorded as verified consensus and 43.6% (41 of 94) on those recorded as adjudicated.
The most frequent differences involved reference Sociopolitical Anxiety and Personal Emotional Expression posts labeled Philosophical / Advisory by the additional annotator (10 and 8 posts, respectively). Six reference residual posts were labeled SociopoliticalAnxiety, andanothersixwerelabeled Humorous / Ironic Use.

These results measure agreement between one additional annotator and a reference built by reviewers who had seen the first-round labels. They are not directly comparable with first-round $\kappa$=0.378, which measures agreement between two independent annotators across all 1,000 posts. The differ-ence between these scores does not demonstrate an improvement in annotation reliability.

| Metric | Value [95% CI] |
|---|---|
| Accuracy | 79.8 [77.2, 82.3] |
| Macro-F1 | .784 [.757, .811] |
| Weighted F1 | .789 |
| Cohen's $\kappa$ | .749 [.717, .778] |
| Majority-class baseline | 27.1 |
| Random baseline (prior-wtd.) | 19.2 |

Table 3: Fanar-C-2-27B against the reference, $n$=1000. Accuracy and baselines in %; percentile bootstrap intervals, $B$=2000, seed 42.

# 6 Model Evaluation

## 6.1 Setup

We evaluated Fanar-C-2-27B,[1] an Arabic-centric LLM, on all 1,000 posts. The prompt lists the six definitions, the tie-break order and the repost rule, and asks for one label and a confidence rating without an explanation. Six hand-written demonstrations, one per category, were checked to be absent from the evaluation set. All 1,000 outputs parsed to a valid label. Appendix D gives the full prompt and decoding settings.

**Scope of the evaluation.** The evaluation characterizes one model under one English-instruction prompt with Arabic examples. We did not run fine-tuned discriminative baselines (e.g., AraBERT, MARBERT or MentalMARBERT), Arabic-instruction prompts, or alternative prompting paradigms. Given evidence that zero-shot LLMs often trail fine-tuned classifiers on social science classification (Mu et al., 2023), we make no claim that Fanar's scores are competitive with supervised models. The released split (Section 8) is intended to make such comparisons straightforward.

## 6.2 Results

Appendix A defines every metric. Table 3 reports overall performance against the reference.

**Readingtheagreementfigurestogether.**

Modelreference $\kappa$ is 0.749 [0.717, 0.778] and first-round human$\kappa$was0.378 [0.338, 0.417], both on the same 1,000 posts. They are not comparable: the human figure measures agreement between two independent annotators, while the model figure measures agreement with a reference built by review and consensus. The gap is not evidence that the model outperforms human annotators. The 200-post validation (Section 5.5) compares one additional annotator with the reference on a subsample and is also not directly comparable with first-round $\kappa$.

| Category | $n$ | P | R | F1 |
|---|---|---|---|---|
| Personal Emotional | 220 | .781 | .859 | .818 |
| Sociopolitical Anxiety | 271 | .803 | .919 | .857 |
| Philosophical / Advisory | 173 | .824 | .728 | .773 |
| Religious / Spiritual | 72 | .833 | .903 | .867 |
| Humorous / Ironic | 114 | .713 | .895 | .794 |
| Not interpretable | 150 | .905 | .447 | .598 |
| *Macro average* | | .810 | .792 | .784 |

Table 4: Per-category precision, recall and F1. $n$ is the number of reference posts in the category.

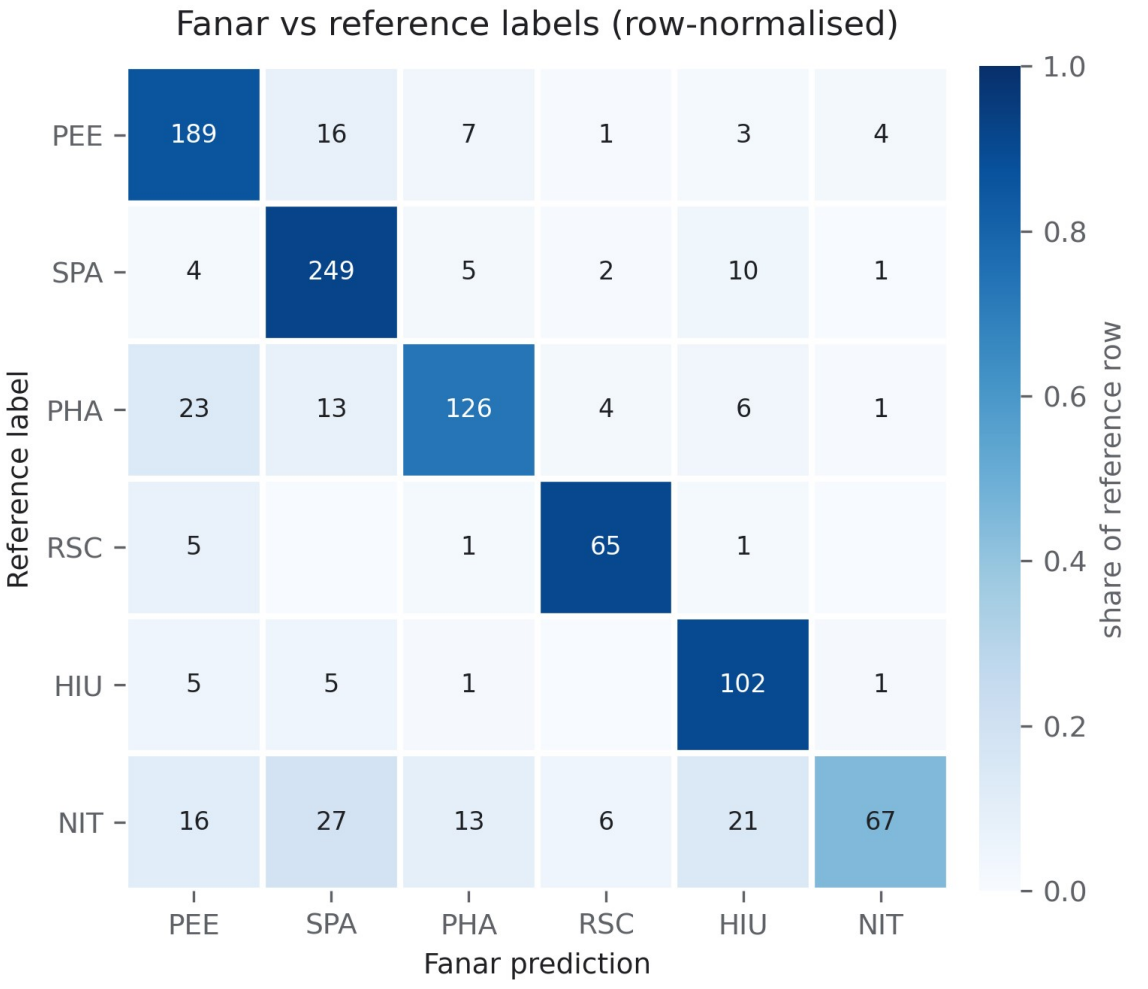

Figure 1: Confusion matrix against the reference, rownormalized; cell values are counts. Every row except the residual category is strongly diagonal.

**6.3 Error analysis: the residual category** Five categories reach F1 between 0.77 and 0.87 (Table 4). The residual category reaches 0.598, combining high precision (0.905) with low recall (0.447): of its 150 reference posts, the model gave 83 a substantive label, and these 83 posts account for 41.1% of all 202 errors.

Figure 1 shows the pattern: every row except the residual one is strongly diagonal. On the 850 posts whose reference label is substantive, accuracy is 86.0% [83.5, 88.4] and macro-F1

[1] Fanar API: https://api.fanar.qa/v1. Model: Fanar-C2-27B.

0.864, against a 31.9% majority baseline. This conditional diagnostic selects posts by their reference label and does not estimate deployment performance.

Three factors limit interpretation. First, the residual category mixes advertisements, readable outofscope posts, insufficient context and uninterpretable text, so its aggregate recall estimates none of these groups separately. Second, the executed prompt lists advertisements as residual but also forbids the residual label for readable posts (Appendix D); a readable advertisement therefore receives contradictory instructions, and without a corrected-prompt comparison the errors cannot be attributed to model capabilityalone. Third, predictingtheresiduallabel is not the same as abstaining. The results support manual checking of residual cases in this configuration, not a general conclusion about LLM-based filtering.

| Category | $n$ | % |
|---|---|---|
| Sociopolitical Anxiety | 271 | 27.1 |
| Personal Emotional Expr. | 220 | 22.0 |
| Philosophical / Advisory | 173 | 17.3 |
| Not interpretable | 150 | 15.0 |
| Humorous / Ironic | 114 | 11.4 |
| Religious / Spiritual | 72 | 7.2 |

Table 5: Reference label distribution, $n$=1000.

**Sensitivity to the reference version.** Keeping predictions fixed, we compared the final reference with the superseded version in which the 511 agreed posts had not been reviewed. Under the supersededreferencetheresidual category held244posts, 79 residual posts were predicted as

Sociopolitical Anxiety, accuracy was 73.1% and $\kappa$ 0.671. Under the final reference that cell falls to 27, total errors fall from 269 to 202, residual recall rises from 0.283 to 0.447, and the residual share of errors falls from 65.1% to 41.1%. These figures are referenceversion diagnostics, not benchmark scores. They show that measured performance on this boundary depends heavily on how the reference is built; they do not show whether the original disagreements came from ambiguous instructions, annotation errors or differing interpretations.

## 7 What Does the Query Retrieve?

Table 5 shows the distribution of reference labels. Sociopolitical Anxiety is the largest category (27.1%), followed by Personal Emotional Expression (22.0%) and Philosophical / Advisory (17.3%); the residual category holds 15.0%.

These are assignment proportions under the tiebreak order, not retrieval precision estimates. Because humour and religious framing rank above first-person expression, a post in which the author voices their own anxiety through prayer is labeled Religious / Spiritual Coping. The 22.0% is therefore a lower bound on posts that contain a firstperson anxiety signal. Estimating retrieval precision for personal anxiety requires a second, multilabel layer that marks first-person anxiety independently of the primary function; we have not annotated that layer, and the release format reserves a field for it.

Even with these caveats, a keyword sample whose largest category is collective concern, and in which more than a quarter of posts are not personal expressions of any kind, should be described by its discourse composition before it is used as an anxiety corpus. Appendix E reports a secondary, descriptive association between discourse type and estimated reach.

## 8 Dataset Release

We will release the following under the provider's licensing terms and our ethics approval:

- the annotation guidelines, the residual-subtype scheme, the executed prompt, and the evaluation code;
- for each post: both first-round labels, the reference label, its provenance (reviewed disagreement or reviewed agreement; matched annotator 1, annotator 2 or neither), the residual sub-reason where recorded, the model prediction, Content Type and Reach;
- a fixed split of 600 / 100 / 300 posts (train / development / test), stratified by reference label with seed 42, so that supervised baselines can be trained and compared on a common test
set;

- a SHA-256 hash of each post identifier, so that researchers holding licensed access to the same posts can verify membership without the identifiers being redistributed.

Post texts will be shared on request for research use, subject to the provider's terms and privacy requirements. Usernames and other identifying information were removed from all examples in this paper.

## 9 Conclusion

We release a benchmark of 1,000 Arabic قلق posts labeled for discourse function. Independent firstround agreement was $\kappa$=0.378. Reviewing all disagreements and all agreements under the same guidelines changed 120 agreed labels, which shows that agreements deserve the same scrutiny as disagreements; the resulting reference is a consensus reference, and on a random 200-post sample an additional annotator agreed with it on 60.5% of posts ($\kappa$=0.517). Fanar-C-2-27B reaches 79.8% accuracy and macro-F1 0.784 against it, with the residual category producing 41.1% of errors; given that category's mixed content and the conflicting prompt instructions, this weakness cannot be attributed to the model alone. Sociopolitical Anxiety (27.1%) outnumbers Personal Emotional Expression (22.0%) as a primary function, which is a statement about this sample's composition and not about the prevalence of anxiety. The released split, provenance fields and residual subtypes are intended to support an independent re-annotation round, supervised baselines, and corrected-prompt comparisons. **Limitations**

**Scope.** One query term, one platform, one provider and a seven-month window. The corpus samples posts containing one polysemous word, not anxiety expression, and supports no conclusions about Arabic-speaking populations, prevalence or mental states.

**Sampling.** The pool was shaped by two genderfiltered exports, removal of 47% exact duplicates, and stratification by content type to prededuplication proportions. Category proportions are properties of this sample and were not validated against an independent random draw.

**Reliability of the reference.** The final labels come from consensus review by two authors under the same guidelines, not from a second independent round. Anchoring and drift in boundary decisions cannot be ruled out, and first-round $\kappa$=0.378 remains the only agreement estimate between two independent annotators. The 200-post validation ($\kappa$=0.517) compares one additional annotator with a reference whose reviewers had seen the first-round labels. It covers a random fifth of the posts and does not assess the residual subtypes. **Residual category.** The category combines advertisements, readable out-of-scope posts, insufficient context and uninterpretable text. Sub-reasons were recorded only for the 95 agreed posts that ended with this label (uninterpretable 42, out of scope 27, advertisements 15, insufficient context 11); four further posts carried an insufficientcontext note but received a substantive label. No sub-reasons exist for the 489 disputed posts, so a subtype-level error analysis over all 150 residual posts is not possible with the current data.

**Prompt ambiguity.** The prompt both assigns advertisements to the residual label and forbids that label for readable posts. Its effect on residual recall cannot be isolated without a corrected-prompt run. **Models and baselines.** We evaluate one LLM under one prompt in one instruction language and report no fine-tuned baselines. Results do not generalize to other models, prompt languages, demonstration sets or prompting paradigms.

**Single-label design.** The tie-break order hides first-person anxiety when humour or prayer is present, so the share of posts with a personal anxiety signal is not estimated.

**Text completeness, dialect and register.** Some posts are truncated mid-sentence; severe truncation leads to the residual label, while milder truncation may affect borderline cases. Truncation prevalence and its link to residual labels were not measured, and dialect and register were not annotated.

**Account-level confounds.** Meltwater exports no follower counts, verification status or account type, so the discourse-reach association is descriptive and may reflect who posts each type of content.

**Size.** One thousand posts support the estimates here, with the smallest category at $n$=72; percategory figures should be read with their support counts, and we do not report per-category intervals. **Ethical Considerations**

Public Twitter/X posts were obtained through Meltwater, a licensed social media monitoring platform. The data collection protocol was reviewed and approved by an Institutional Review Board; approval details will be given in the camera-ready version. We did not contact any user or request additional data.

**Privacy.** We release no usernames, user identifiers, account metadata or raw post identifiers. Identifying information was removed from all examples in this paper, and the prompt demonstrations were written by hand.

**Sensitive content.** The corpus contains posts about anxiety and psychological distress. We diagnose no individual and report findings only at the group level. Our categories describe how language is used, not the mental state of any author: a post labeled Personal Emotional Expression shows first-person framing, not a clinical condition.

**Annotator welfare.** Annotators were told about the content domain before starting, were free to stop at any time, and were advised to spread the work across sessions.

**AI assistants.** AI writing assistants were used for language refinement and literature search. All analyses, annotation pipelines and results were produced by the authors; no AI tool generated or modified experimental results. Fanar-C-2-27B is an object of study, not an authoring tool.

## A Metrics

Let $N$ be the number of evaluated posts, $\mathrm{L}$ the set of labels included in macro-averaging, $g_i$ and $p_i$ the reference and predicted labels, and $n_\ell$ the reference support of category $\ell$. For the main evaluation, $N$=1000 and $|\mathrm{L}|$=6. The conditional diagnostic retains posts whose reference label is substantive

($N$=850); predictions are unchanged, so a residual prediction on a retained post is an error and a false negative for its reference category, and macro-F1 averages the five substantive categories. **Agreement.** Cohen's $\kappa$ corrects observed agreement $p_o$ for chance agreement $p_e$:

$$\kappa = \frac{p_o - p_e}{1 - p_e}, \qquad p_o = \frac{1}{N}\sum_i \mathbf{1}[g_i = p_i], \quad (1)$$

$$p_e = \sum_{\ell \in \mathcal{L}} \frac{|\{i : g_i{=}\ell\}|}{N} \cdot \frac{|\{i : p_i{=}\ell\}|}{N}. \quad (2)$$

We apply Eq. 1 between annotators (§5), between the validation annotator and the reference (§5.5), and between model and reference (§6.2); percategory $\kappa$ uses the one-vs-rest binarization of $\ell$.

**Classification.** With $TP_\ell$, $FP_\ell$, $FN_\ell$ the usual counts,

$$P_\ell = \frac{TP_\ell}{TP_\ell + FP_\ell}, \qquad R_\ell = \frac{TP_\ell}{TP_\ell + FN_\ell}, \quad (3)$$

$$F_\ell = \frac{2P_\ell R_\ell}{P_\ell + R_\ell}, \qquad F^{\text{mac}} = \frac{1}{|\mathcal{L}|}\sum_\ell F_\ell, \quad (4)$$

$$F^{\text{wgt}} = \frac{1}{N}\sum_\ell n_\ell F_\ell. \quad (5)$$

We lead with $F^{\text{mac}}$, which weights categories equally. A quantity with a zero denominator is defined as 0, and $F_\ell$=0 when $P_\ell$+$R_\ell$=0; no category in our results has a zero denominator.

**Baselines.** The majority baseline always predicts the most frequent reference category; drawing predictions from the reference distribution gives expected accuracy $\text{Acc}_{\text{rand}} = \sum_\ell (n_\ell/N)^2$.

**Intervals.** For each of $B$=2000 resamples (seed 42) we draw $N$ post indices with replacement, keep the paired labels, and recompute the statistic; endpoints are the 2.5th and 97.5th percentiles. Intervals are conditional on the recorded labels and exclude uncertainty from alternative adjudication decisions.

## B Annotation Codebook

The category definitions in Section 4 were used for both annotation and review.

**Tie-break order.** Applied in sequence, stopping at the first that applies: (1) joke or sarcasm → Humorous / Ironic Use; (2) prayer, verse or hadith → Religious / Spiritual Coping; (3) advice or reflection addressed outward → Philosophical / Advisory; (4) the author's own current state → Personal Emotional Expression; (5) public events or a news report → Sociopolitical Anxiety; (6) otherwise → Not interpretable.

**Additional rules.** Posts beginning RT @ or QT @ are labeled as the visible text stands. Advertisements for medication, applications or books are Not interpretable. A post consisting only of قلق, only آمين, or a single hashtag is Not interpretable. Laughter markers (ههه, emoji) indicate Humorous / Ironic Use even when the subject is political. A post truncated mid-sentence is labeled if its function is recoverable.

**Review.** The review applied the same definitions andtie-breakrulesanddidnotintroduceanewcodebook version. Appendix D preserves the executed prompt with its conflicting residual instructions.

## C Residual Subtype Scheme

Table 6 defines four residual subtypes with an ordered decision procedure. It applies only after the tie-break order in Appendix B has failed to assign a substantive label, and it is designed so that a readable post never lands in the uninterpretable subtype. The subtypes match the sub-reasons recorded during review of the agreed posts, whose counts are shown; the scheme was not available to the firstround annotators and is offered for future annotation, re-annotation and subtype-level evaluation. A corrected prompt should replace the single residual definition in Appendix D with these four rules, removing the instruction that forbids the residual label for readable posts.

| Subtype | Decision rule (apply in order) | Typical cases | Agreed $n$ |
|---|---|---|---|
| R1 Advertisement | The post promotes a product, service, app, book, course or account. | Medication or therapy-app promotion; book sales; paid course announcements. | 15 |
| R2 Out of scope | The post is readable and its function is clear, but it fits none of the five substantive definitions. | Readable posts whose topic is unrelated to worry or concern; non-promotional spam. | 27 |
| R3 Insufficient context | The post is readable, but too little text remains to determine its function. | Only قلق, only آمين, a single hashtag; replies whose meaning depends on an unseen post; truncation that removes the predicate. | 11 |
| R4 Uninterpretable | The text cannot be read as meaningful Arabic. | Empty text; non-Arabic text; word salad; encoding damage. | 42 |

Table 6: Residual subtypes. "Agreed $n$" counts the 95 initially agreed posts whose final label was residual; no sub-reasons were recorded for the remaining 55 residual posts, which came from the 489 disagreements.

## D Model Prompt and Settings

The text below reproduces the executed task instructions. The notebook built the per-post messages and output-format instructions from the task settings. Instructions were in English with Arabic examples; no Arabic-instruction variant was tested.

Classify how the word قلق (or its inflections: قلقان ، قلقه ،القلق ،مقلق ،قلقة) is used in an Arabic social media post. Assign exactly one label.

**LABEL DEFINITIONS**

**Personal Emotional Expression** The author expresses their own anxiety, distress, worry or unease, in the first person, about their own state. Example: مش قادر أنام من القلق، بقالي أسبوع كده

**Sociopolitical Anxiety**

Worry or concern about political events, war, the economy, or a collective situation — whether felt by the author, attributed to someone else, or reported. This INCLUDES news reports and diplomatic formulas such as أعرب عن قلقه، قلق دولي، يتابع

الوضع.

بقلققلق دولي متزايد من

التصعيد في المنطقة بعد

Example: بيان

أمس

**Philosophical / Advisory**

Advice, reflection, or general wisdom about anxiety, addressed to others or to people in general, rather than a report of the author's own current state. Example: إذا زاد عليك القلق فاعلم أن أكثر ما تخافه لن

يحدث

**Religious / Spiritual Coping** A du'a, prayer, Quranic verse, hadith, or invocation of God in connection with anxiety. Example: اللهم اصرف عنا القلق واملأ قلو بنا طمأنينة

**Humorous / Ironic Use**

A joke, meme, sarcastic or mocking remark. Laughter markers (ههه,) and moc😂ry🤣f stock phrases belong here even when the subject is political. Example: وظيفة أحلامي: أعبّر عن قلقي البالغ وأستلم راتب

بالدولار هههه

**Not interpretable**

ONLY: empty text, text not in Arabic, random or word-salad content, advertisements, or text truncated so badly that its meaning cannot be recovered. If you can read and understand the post, do NOT use this label — choose one of the five above. Example: قلق شدد مخفور جرجر نقه متبر بر

**TIE-BREAK ORDER**

When two labels both seem to fit, apply these in order and stop at the first that applies:

1. Is it a joke or sarcasm? → Humorous / Ironic Use
2. Is it a prayer, verse or hadith? → Religious / Spiritual Coping
3. Is it advice or reflection addressed outward?→ Philosophical / Advisory
4. Is it about the author's own current state? → Personal Emotional Expression
5. Is it about public events, or is it a news report?→ Sociopolitical Anxiety
6. Otherwise → Not interpretable

**ADDITIONAL RULES**

– If the post begins with RT @ or QT @, classify the visible text as it stands. Do not attempt to recover the original post. – Advertisements for medication, apps or books about anxiety → Not interpretable – Apostconsistingonlyof قلق, only آمين, or a single hashtag → Not interpretable – A post may be truncated mid-sentence. If you can still tell what it is doing, classify it; use Not interpretable only when you cannot.

**Task configuration.** The notebook settings were post_language="Arabic", multi_label=False,

include_confidence=True and include_explanation=False, with Not interpretable as the fallback label.

**Settings.** Model Fanar-C-2-27B; temperature 0; top-$p$ 1.0; maximum 80 new tokens; one greedy run. Outputs were matched against the six category names; all 1,000 posts yielded a valid label on the first attempt, with no retries or refusals. The confidence field returned high for every post, so it carries no information and is not used. The run configuration, model inputs and raw predictions accompany the released code.

## E Discourse Type and Visibility

Reach is right-skewed, so we compare medians with rank-based tests: Kruskal–Wallis $H$ across the six categories, and Mann–Whitney $U$ with Cliff's $\delta$ between two. With group 1 the first-named category,

$$\delta = \frac{2U}{n_1 n_2} - 1 \in [-1, 1], \tag{6}$$

so $\delta > 0$ means group 1 has higher ranks. Under Romano et al. (2006)'s thresholds, $|\delta| < 0.147$ is negligible and $< 0.33$ small.

All figures use the reference labels. Reach is available for 975 of the 1,000 posts. Median reach differs across categories ($H$=53.6, $p$=$2.5\times10^{-10}$): Sociopolitical Anxiety 530 ($n$=271) and Humorous / Ironic Use 688 ($n$=111) against Personal Emotional Expression 179 ($n$=215) and the residual category 121 ($n$=137). Sociopolitical Anxiety versus Personal Emotional Expression gives $\delta$ = +0.254 ($p$=$1.6\times10^{-6}$), a small effect.

Among the 667 posts with available Reach that Content Type does not mark as reposts, the association persists: medians 532 ($n$=128) against 144.5 ($n$=176), $\delta$ = +0.260, $p$=$1.1\times10^{-4}$; $H$=47.5, $p$=$4.4\times10^{-9}$ across all six.

We state this narrowly. Excluding reposts does not establish independence from reposting, since the remaining posts still circulate through accounts of very different sizes, and without account-level fields discourse effects cannot be separated from account effects. We make no claim about platform amplification. The Reach and Content Type fields and the script 04_visibility.py are included in the release.

## F Reproducibility

All intervals are percentile bootstrap intervals ($B$=2000, seed 42) for accuracy, macro-F1 and Cohen's $\kappa$, conditional on the recorded labels; $\kappa$ bands follow Landis and Koch (1977). The evaluation pipelinechecksfourpreconditions beforeproducing final results: exactly 1,000 unique post identifiers; no missing reference label; all labels within the codebook; and no duplicated label column among the adjudicated rows. The last check guards against copying a column; it does not demonstrate independent judgment. Runs failing any check are marked exploratory; all four passed for the results reported here.